\pdfoutput=1
\documentclass{article}

\usepackage{PRIMEarxiv}
\usepackage[utf8]{inputenc}
\usepackage[T1]{fontenc}
\usepackage{amsmath,amssymb}
\usepackage{graphicx}
\usepackage{subcaption}
\usepackage{booktabs,array}
\usepackage{enumitem}
\usepackage[table,dvipsnames]{xcolor}
\usepackage{microtype}
\usepackage{tikz}
\usetikzlibrary{spy,calc}
\usepackage{cite}
\usepackage{xurl}
\usepackage[hidelinks,breaklinks,colorlinks=true,linkcolor=blue, citecolor=blue, urlcolor=blue]{hyperref}
\usepackage[capitalise]{cleveref}

\newcommand{\parcelstowrepotxt}{https://github.com/coenwerem/parcelstow}
\newcommand{\parcelstowrepo}{\url{\parcelstowrepotxt}}

\definecolor{oursmint}{HTML}{E2F1EA}
\definecolor{losssalmon}{HTML}{F9E8E4}

\newcommand{\FC}{\varepsilon}

\crefname{section}{Sec.}{Secs.}
\Crefname{section}{Sec.}{Secs.}
\crefname{figure}{Fig.}{Figs.}
\Crefname{figure}{Fig.}{Figs.}
\crefname{table}{Table}{Tables}
\Crefname{table}{Table}{Tables}
\crefname{appendix}{Appendix}{Appendices}
\Crefname{appendix}{Appendix}{Appendices}
\crefname{equation}{Eq.}{Eqs.}
\Crefname{equation}{Eq.}{Eqs.}

\title{\normalfont \Large {\textbf{Does Imitation Learning Preserve Temporal Robustness in Dexterous Manipulation?}\\An Expert--Learner Comparison Across Task Execution Speeds}}

\author{
  Clinton Enwerem, John S. Baras, Calin Belta \\
  Institute for Systems Research \\
  University of Maryland \\
  College Park, MD, U.S.A.\\
  \texttt{\{enwerem, baras, calin\}@umd.edu} \\
}

\begin{document}
\raggedbottom
\maketitle

\begin{abstract}
\noindent
Dexterous manipulation policies learned by imitation are typically evaluated for robustness to variation in scenes, objects, or instructions, but their performance across task execution speeds is less often examined. This leaves open how much temporal robustness a learner retains relative to the expert it imitates. We compare an expert and learner under the same task conditions, initial-condition draws, and speedup factors. We instantiate the evaluation in ParcelStow, a contact-rich task in which the robot acquires, reorients, and inserts a parcel. The demonstrations span the speedup range for the manipulation phases after parcel acquisition. A scripted expert and an Action Chunking with Transformers (ACT) policy trained from the expert's demonstrations both achieve 100\% task success at nominal speed. Their success rates diverge within the demonstrated range: at its maximum, expert success is 84\% and ACT success is 53\%. Two ACT policies with different parameter initializations show similar degradation, decreasing by 34 and 48 percentage points from nominal speed to the maximum demonstrated speed, compared with 16 points for the expert. Stage-level analysis shows that 35 of ACT's 47 failures at the maximum demonstrated speed are insertion misalignments. Under the relative-motion handoff, every ACT acquisition retains the parcel through reorientation and transfer in free space, but only 64\% complete the overall task, compared with 95\% after expert acquisition. Across all evaluated policies and speeds, none of the 414 acquisitions without force closure completes the task. Equal nominal task success therefore does not imply preservation of expert performance across execution speeds. Code, data, and evaluation scripts are available at \parcelstowrepo.
\end{abstract}
\section{Introduction}\label{sec:intro}
Imitation learning transfers demonstrated behavior to a policy, but nominal task success alone does not establish preservation of expert performance across execution speeds. In contact-rich manipulation, changing execution time alters velocities, accelerations, tracking demands, and contact transients and can change task success. An imitation policy may therefore reproduce nominal expert performance without reproducing the expert's dependence on execution speed. Existing benchmarks primarily vary tasks, scenes, objects, or instructions rather than execution speed~\cite{yu2019metaworld,james2019rlbench,mees2021calvin,liu2023libero,nasiriany2024robocasa}.

We vary task execution speed and compare an expert policy with a learned imitation policy at the same speeds. Hereafter, we refer to these policies as the \emph{expert} and \emph{learner}. Recent methods have accelerated demonstrations, retimed predicted actions, or adapted learned policies for faster execution \cite{guo2025demospeedup,arachchige2025sail,nam2026speedaug}. These approaches primarily evaluate execution beyond the demonstration speed. We instead compare how expert and learner success rates change at matched speeds, including speeds in the support of the demonstrated speed distribution.

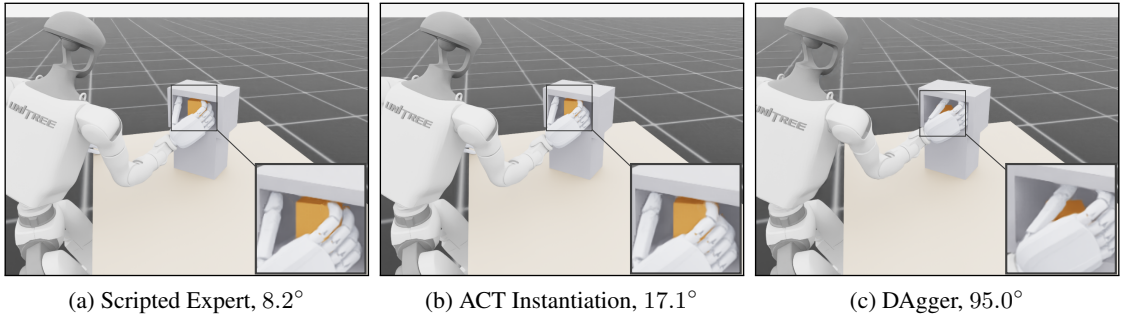
\begin{figure}[t]
\centering
\setlength{\tabcolsep}{2pt}
\setlength{\fboxsep}{0pt}
\setlength{\fboxrule}{0.5pt}
\newcommand{\relpt}[3]{($($(#1.south west)!#2!(#1.south east)$)!#3!($(#1.north west)!#2!(#1.north east)$)$)}
\newcommand{\teaserspy}[3]{
\begin{tikzpicture}[spy using outlines={rectangle, draw=black!75, semithick, magnification=2.4, size=1.45cm, connect spies}]
\node[anchor=south west, inner sep=0, outer sep=0] (img) {\fbox{\includegraphics[width=0.29\textwidth]{figures/stills/#1}}};
\spy on \relpt{img}{#2}{#3} in node[anchor=south east] at \relpt{img}{0.99}{0.01};
\end{tikzpicture}}
\begin{tabular}{ccc}
\teaserspy{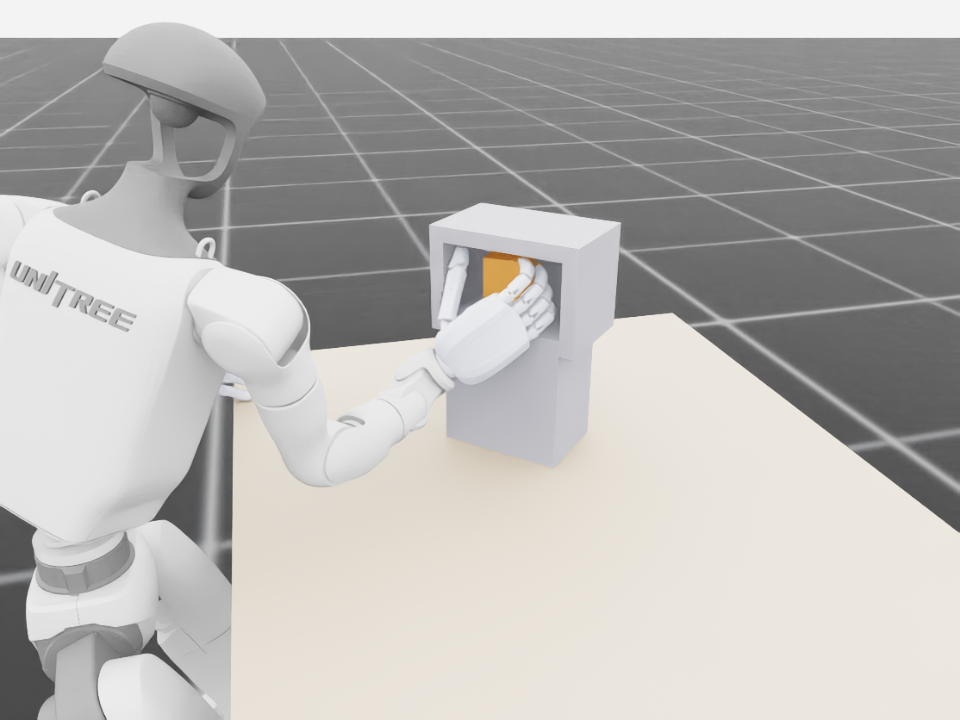}{0.520}{0.615} &
\teaserspy{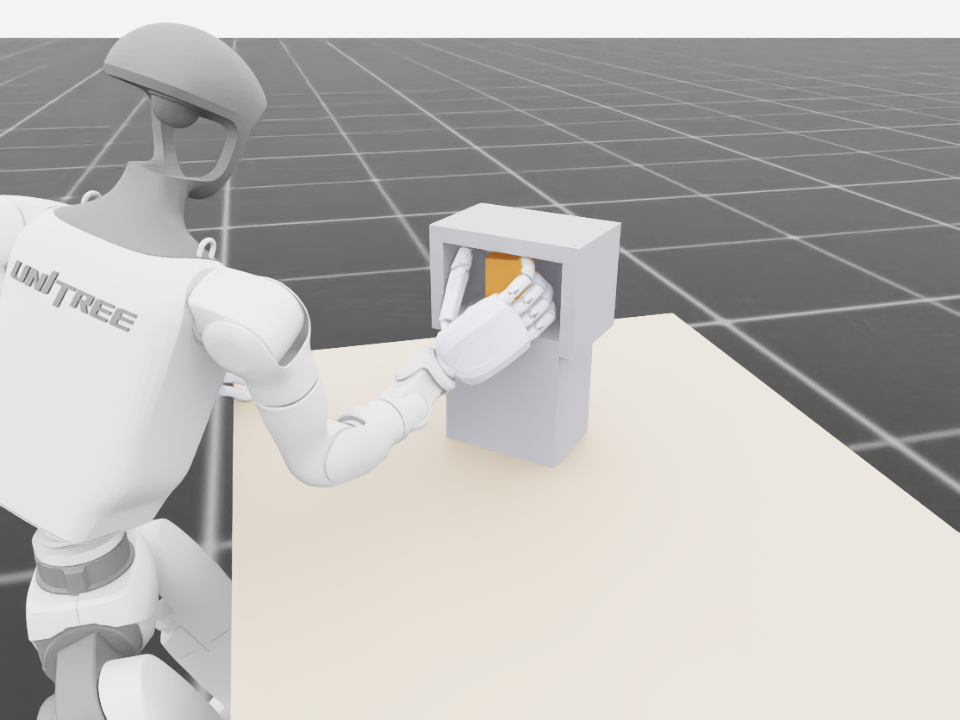}{0.520}{0.615} &
\teaserspy{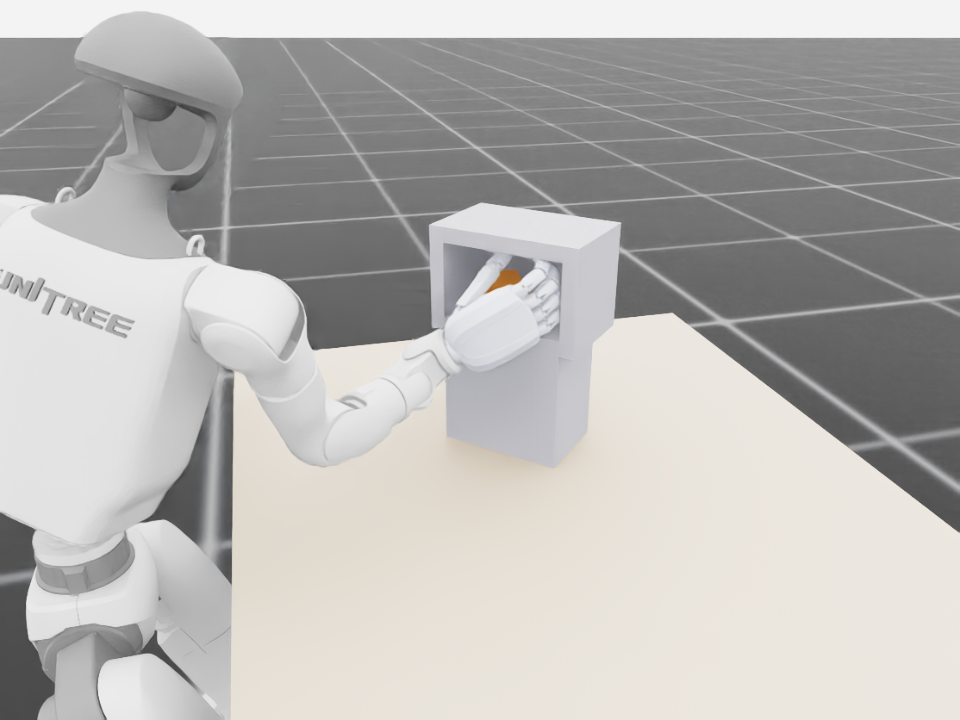}{0.515}{0.595}\\[1pt]
\small (a) Scripted Expert, $8.2^\circ$ & \small (b) ACT Instantiation, $17.1^\circ$ & \small (c) DAgger, $95.0^\circ$
\end{tabular}
\caption{\textbf{Dexterous Insertion at the Maximum Demonstrated Speed.} Representative terminal states of the parcel acquisition, reorientation, and insertion task (\Cref{sec:setup}), with magnified views of the receptacle interior. Each subcaption reports the final parcel orientation error; the settling tolerance is $10^\circ$. (a) The expert places the parcel flush and upright, within tolerance. (b) ACT-A leaves the parcel rotated and outside tolerance. (c) DAgger topples the parcel onto its side.}
\label{fig:teaser}
\end{figure}

We evaluate the expert and learner under the same task conditions and initial-condition draws at each speed. Their task-success curves characterize each policy's dependence on execution speed, and their signed difference gives the expert--learner success-rate difference at a common speed. The comparison itself does not depend on a particular robot embodiment, policy architecture, or manipulation task.

To conduct this comparison, we present ParcelStow, a contact-rich manipulation benchmark with modular interfaces for the robot embodiment, policy, and task. We evaluate Action Chunking with Transformers (ACT)~\cite{zhao2023act}, DAgger~\cite{ross2011dagger}, and Diffusion Policy~\cite{chi2023diffusion} on a parcel acquisition, reorientation, and insertion task using a fixed-base humanoid. Among the three learners, ACT matches the expert's nominal 100\% success rate yet exhibits a larger decrease in success across the demonstrated speed range. Stage outcomes place most ACT failures during insertion, and a relative-motion handoff rules out loss of parcel retention during reorientation and free-space transfer under the shared hand motion. \Cref{fig:teaser} depicts representative terminal states at the maximum demonstrated speed. We summarize our contributions next.
\paragraph{Contributions}
\begin{enumerate}[label=\roman*.,leftmargin=*,noitemsep]
    \item \textbf{Comparison of Expert and Learner Task Success:} We formulate a controlled comparison of expert and learner task success rates across execution speeds under shared task conditions, success criteria, and draws of initial conditions (\Cref{sec:problem}).
    \item \textbf{Controlled Benchmark:} We introduce ParcelStow, a contact-rich manipulation benchmark for this comparison with a demonstrated speed range, a common action interface, and a specified execution-speed range (\Cref{sec:exps}).
    \item \textbf{Empirical Study of Temporal Sensitivity:} We compare the success curves of a scripted expert and three learners across the demonstrated speed range (\Cref{sec:results}).
    \item \textbf{Policy Failure Diagnostics:} We combine task outcomes by stage, arm joint velocity utilization, a relative-motion handoff, hand--parcel relative motion, and force closure at acquisition to locate and characterize policy failures without changing the success criterion for the overall task (\Cref{ssec:diagnostics}).
\end{enumerate}

\subsection{Related Work}\label{ssec:related}
\subsubsection{Imitation and Generalist Manipulation}\label{sssec:imitgenman}
Offline imitation performance varies with demonstration coverage, model design, action representation, and the difference between training and evaluation state distributions~\cite{mandlekar2021robomimic}. DAgger addresses the sequential train--test mismatch by aggregating expert labels on states induced by the learner~\cite{ross2011dagger}. BeT and Diffusion Policy model multimodal action distributions~\cite{shafiullah2022bet,chi2023diffusion}, PerAct extends behavior cloning to multi-task manipulation through voxelized representations~\cite{shridhar2022peract}, and ACT predicts multi-step action chunks with temporal ensembling~\cite{zhao2023act}. Comparative studies have also examined imitation-learning design choices~\cite{jia2025xil}, in-context imitation~\cite{vosylius2025instantpolicy}, cross-embodiment scaling~\cite{openx2023,octo2024,black2024pi0}, and discrepancies between demonstrated and executed motion~\cite{kedia2024mismatched}. These studies establish that imitation performance depends on both policy design and the conditions represented in the demonstration data.

Several recent methods explicitly accelerate policy execution. DemoSpeedup trains policies on demonstrations downsampled at segment-specific rates \cite{guo2025demospeedup}, while SAIL combines temporally consistent inference, controller-invariant targets, adaptive speed modulation, and latency-aware action scheduling \cite{arachchige2025sail}. SpeedAug uses reinforcement learning to adapt a policy initialized from speed-augmented demonstrations \cite{nam2026speedaug}. Other chunk-based methods reduce inference delays by asynchronously generating future actions~\cite{black2025rtc} or adapting execution horizons~\cite{nie2026pace}. These works evaluate modified policies under accelerated execution. ParcelStow instead compares a learner directly with its demonstration expert at matched execution speeds, including speeds within the demonstrated range.

\subsubsection{Dexterous and Dynamic Manipulation}\label{sssec:dexdynman}
Shorter execution times increase velocities, accelerations, and transient contact loads in dexterous manipulation~\cite{rajeswaran2017dexterous,andrychowicz2018inhand,openai2019rubik,petrenko2023dexpbt,sleimanGuidedReinforcementLearning2024}. Recent systems train grasps for dynamic motion~\cite{zhang2025robustdexgrasp}, convert static demonstrations into dynamic manipulation~\cite{liao2026dynamicmanip}, or select grasps using the downstream motion they must support~\cite{gupta2026grasptoact,enwerem_equidexflow_2026}. These methods seek grasps that remain stable during downstream dynamic motion. Research on grasp quality separately evaluates properties of the grasp wrench space, primarily force and form closure~\cite{murrayMathematicalIntroductionRobotic2017a}, and task execution through analysis or simulation~\cite{ferrari1992planning,roaGraspQualityMeasures2015,li2023frogger,wang_dexgraspnet_2023,sundermeyer2021contact,morrison2018ggcnn,enwerem2026firmgrasp,enweremVariationalNeuralBeliefParameterizations2026a}. ParcelStow assumes task-compatible grasps are available and uses the force closure margin~\cite{murrayMathematicalIntroductionRobotic2017a} only to characterize realized hand--object contacts.

\subsubsection{Policy Evaluation and Controlled Perturbations}\label{sssec:rbtpleval}
Robot learning benchmarks specify tasks, demonstrations, and controlled tests of generalization \cite{yu2019metaworld,james2019rlbench,mees2021calvin,liu2023libero,zhu2020robosuite,gu2023maniskill2,tao2024maniskill3,nasiriany2024robocasa}. Recent work also studies evaluation procedures. Li et al.~\cite{li2025simpler} compare relative policy performance in simulation and hardware, Zhou et al.~\cite{zhou2025autoeval} automate repeated hardware trials, Pumacay et al.~\cite{pumacay2024colosseum} measure success under fourteen environmental perturbations, Sagar et al.~\cite{sagar2024robomd} search structured environment variations, and Jiang et al.~\cite{jiang2026benchmarkaudit} examine the representativeness of fixed benchmark scores across a broader task set. These evaluations vary scenes, objects, environmental conditions, or task specifications. ParcelStow instead keeps the scene, geometry, and task goal fixed while shortening scheduled execution time. It compares each learner with its demonstration expert at the same execution speeds and determines how their success rates differ as speed changes.

\section{Problem Formulation}\label{sec:problem}
We consider a manipulation task with physical state $x_t\in\mathcal X$, action $u_t\in\mathcal U$, and ordered task phases $\mathcal K=\{k_1,\ldots,k_K\}$. The initial condition $\omega\in\Omega$ specifies the randomized components of the initial physical state through $x_0=x_{\mathrm{init}}(\omega)$. Let $P_\Omega$ be their distribution over the initial-condition space $\Omega$. The task also maintains a phase variable $\phi_t$, the previous action $u_{t-1}$, and an execution-speed factor $r$. We collect these variables in the augmented state $z_t=(x_t,\phi_t,u_{t-1},r)\in\mathcal Z$, whose closed-loop transition satisfies $z_{t+1}=F(z_t,u_t)$.

The learner observation map $g:\mathcal Z\to\mathcal O$ produces $o_t=g(z_t)$, and the learner acts according to $u_t=\pi_L(o_t)$ for $\pi_L:\mathcal O\to\mathcal U$. The expert can depend on different components of the augmented state. We represent this dependence by an expert input map $h_E:\mathcal Z\to\mathcal I_E$ and policy $\pi_E:\mathcal I_E\to\mathcal U$, so its action is $u_t=\pi_E(h_E(z_t))$. Thus, $\mathcal I_E$ is the expert input space, whereas $\mathcal O$ is the learner observation space. Both policies act through the same action space, physical dynamics, task geometry, and success criterion.

The speedup factor lies in an evaluation set $\mathcal R_E\subset\mathbb R_{>0}$, and $r_0=1$ denotes nominal execution. Demonstrations are generated over a distribution $P_D$ whose support is the demonstrated speed range $\mathcal R_D\subseteq\mathcal R_E$. Let $\mathcal M\subseteq\mathcal K$ denote the task phases whose durations vary. If phase $k$ has nominal duration $T_k$, its duration at speedup factor $r$ is
\begin{equation}
T_k(r)=
\begin{cases}
T_k/r, & k\in\mathcal M,\\
T_k, & k\notin\mathcal M.
\end{cases}
\label{eq:rate}
\end{equation}
Changing $r$ changes the phase schedule and the number of policy steps $H(r)$ while leaving the task geometry, physical parameters, and success criterion fixed.

For policy $\pi$, speedup factor $r$, and initial condition $\omega$, let
\begin{equation}
\tau(\pi,r,\omega)=\left(z_0,u_0,\ldots,z_{H(r)}\right)
\label{eq:rollout}
\end{equation}
denote the resulting rollout. The binary function $Y(\tau)\in\{0,1\}$ equals one if and only if the rollout satisfies the task success criterion. The policy's task success probability at $r$ is
\begin{equation}
p_\pi(r)=\Pr\!\left[Y\!\left(\tau(\pi,r,\omega)\right)=1\right],
\label{eq:success_probability}
\end{equation}
where the probability is over $\omega\sim P_\Omega$ and any internal randomness of $\pi$. Given $N$ evaluation draws $\omega_1,\ldots,\omega_N$, we estimate $p_\pi(r)$ as
\begin{equation}
\widehat p_{\pi}(r)=\frac{1}{N}\sum_{i=1}^{N}Y\!\left(\tau(\pi,r,\omega_i)\right).
\label{eq:observed_success}
\end{equation}

The learner is trained from an expert demonstration set
\begin{equation}
\mathcal D_E=\left\{\tau_m\right\}_{m=1}^{M}.
\label{eq:demonstrations}
\end{equation}
Each demonstration $\tau_m=\tau(\pi_E,r,\omega)$ uses independent draws $r\sim P_D$ and $\omega\sim P_\Omega$.
For evaluation at a common value of $r$, we apply the same initial-condition draws to the expert and learner. Their signed empirical difference is
\begin{equation}
G_L(r)=\widehat p_{\pi_E}(r)-\widehat p_{\pi_L}(r).
\label{eq:matched_gap}
\end{equation}
Positive values indicate that the expert succeeds more often on the matched evaluation sample. When $\widehat p_{\pi_E}(r_0)=\widehat p_{\pi_L}(r_0)$, $G_L(r)$ for $r\in\mathcal R_E\setminus\{r_0\}$ is the expert--learner success-rate difference at the other evaluated speeds. The full success curves remain important because $G_L(r)$ at one speed does not characterize either policy's dependence on $r$.

Our evaluation tests if nominal expert--learner equality persists at other speeds in the support of the demonstrated speed distribution. The comparison requires a common task, action interface, success criterion, phase-duration scaling, and initial-condition distribution. The expert and learner may use different input spaces. Learners without nominal expert--learner equality provide secondary comparisons of dependence on execution speed.

\section{Methodology}\label{sec:methodology}
We compare an expert and a learner by changing the phase durations in \Cref{eq:rate} and holding the remaining task conditions fixed. At each $r\in\mathcal R_E$, we evaluate both policies on the same ordered initial-condition draws $\omega_1,\ldots,\omega_N$. We preserve the task geometry, physical parameters, success predicates, and distribution $P_\Omega$ across policies and speedup factors. Changes in a policy's success rate across $r$ then characterize its sensitivity to execution speed, and $G_L(r)$ is the signed expert--learner difference at each common speed.

We report $\widehat p_\pi(r)$ for every policy and compute $G_L(r)$ from the matched episode draws. Values of $r\in\mathcal R_D$ evaluate performance within the demonstrated speed range; values of $r\in\mathcal R_E\setminus\mathcal R_D$ evaluate extrapolation beyond that range. A learner satisfying $\widehat p_{\pi_E}(r_0)=\widehat p_{\pi_L}(r_0)$ provides the primary comparison because the expert and learner have equal success rates at nominal speed. Other learners provide secondary comparisons of changes in success across execution speeds.

We supplement the success comparisons with two forms of analysis. Diagnostics computed from recorded rollouts identify the first failed task predicate, quantify arm motion and hand--parcel relative motion, and characterize realized contact sets. In separate handoff rollouts, we replace each policy's commands after parcel acquisition with a shared controller to test parcel retention under common downstream hand motion. Neither analysis changes the speedup factors or the task success function $Y$.

\section{Experiments}\label{sec:exps}
\subsection{Experimental Setup}\label{sec:setup}

\subsubsection{Simulation, Embodiment, and Physical Task}\label{ssec:embtsk}
We instantiate the formulation in ParcelStow, an Isaac Lab task \cite{mittal2023orbit}. The embodiment is a fixed-base Unitree G1 humanoid equipped with a RealHand L6 anthropomorphic right hand. The parcel is an $80\times55\times40$~mm rigid cuboid of mass $0.12$~kg with nominal static and dynamic friction $\mu=0.5$. Its nominal position is $(0.35,\,0,\,0.721)$~m with a $45^\circ$ yaw. An episode acquires the parcel, lifts it clear of the table, rotates it by $90^\circ$ about its width axis, transports it $0.334$~m to a receptacle with an open front, and inserts it along an axis with $10$~mm of clearance per side. The parcel remains a dynamically free rigid body. The scene contains no weld or fixed joint between the hand and parcel and no injected object wrench. Episodes run through the complete phase schedule; no geometric shortcut terminates an episode before settling. Regression tests confirm that the parcel falls when the hand opens, cannot be dragged without contact, collides with the receptacle, and cannot satisfy the insertion predicate outside the usable volume.

Task success is an ordered conjunction of predicates evaluated on simulator state. Acquisition requires the parcel center to rise by at least $20$~mm while the thumb distal phalanx and at least one further distal phalanx each exert more than $1$~N of contact force on the parcel. Lift clearance requires a $60$~mm rise above the table. Reorientation and the pose before insertion require orientation error below $15^\circ$. Insertion requires the parcel center to pass at least $50$~mm beyond the receptacle entrance plane while remaining inside its cross section. Release requires contact force between the distal phalanges and parcel below $0.5$~N for at least $0.1$~s. Settling requires the released parcel to remain inserted for $0.4$~s with linear speed below $0.02$~m/s, angular speed below $0.2$~rad/s, and final orientation error below $10^\circ$. The insertion and settling predicates latch as stage events. The terminal position and orientation criteria must also hold, so an episode can satisfy the insertion predicate and still fail after release. \Cref{app:task} gives the complete specification.

\subsubsection{Observation, Action, and Low-Level Control}\label{ssec:stactctrl}
For the learners, $\mathcal O\subset\mathbb R^{147}$. The observation concatenates full-body joint position and velocity, the previous action, parcel pose in the base frame, right-hand fingertip positions and contact forces, the task phase, and $r$. Observation corruption is disabled during evaluation. Learner actions satisfy $u_t\in\mathcal U\subset\mathbb R^{16}$ and are normalized joint position offsets for the waist, right arm and wrist, and six actuated joints of the right hand. The environment applies
\begin{equation}
q_{\mathrm{target},t}=q_{\mathrm{default}}+0.5u_t
\label{eq:action_map}
\end{equation}
through separate implicit PD drives at a $50$~Hz policy and control interface.

The expert does not consume the learner observation vector. It reads the task phase and current joint configuration from simulator state, computes joint targets by damped least-squares differential inverse kinematics over the waist and right arm, and converts those targets to $u_t$ by inverting \Cref{eq:action_map}. The expert and learners therefore use different information sets but share the same $16$-dimensional action channel and low-level drives. \Cref{app:control} gives the observation slices, joint order, drive gains, simulation step, and inverse kinematics parameters.

\subsubsection{Execution-Speed Conditions}\label{ssec:speedvariation}
\begin{figure}[t]
\centering
\includegraphics[width=0.94\textwidth]{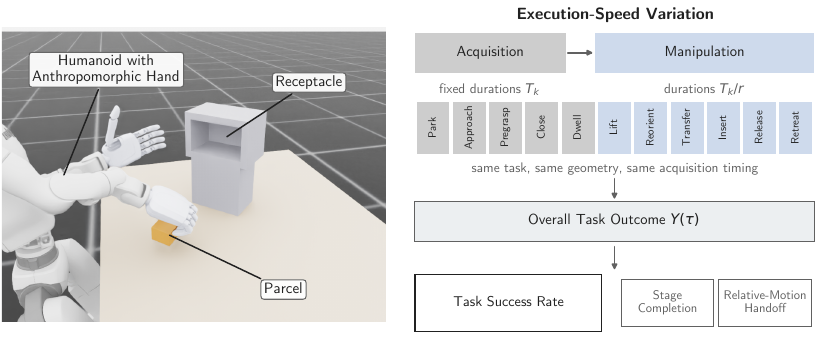}
\caption{\textbf{ParcelStow Task, Embodiment, and Variation in Execution Speed.} Left, the fixed-base humanoid, parcel, and receptacle with an open front. Right, the phase schedule. Acquisition phases retain their nominal durations $T_k$, and the durations of manipulation phases are divided by $r$. Task geometry and acquisition timing remain fixed. Standard evaluation rollouts yield overall and stage-level task outcomes; separate relative-motion handoff rollouts evaluate acquired grasps under shared downstream hand motion.}
\label{fig:system}
\end{figure}

ParcelStow assigns
\begin{equation*}
\begin{aligned}
\mathcal K={}&\{\mathtt{PARK},\mathtt{APPROACH},\mathtt{PREGRASP\_DWELL},\mathtt{CLOSE},\mathtt{GRASP\_DWELL},\mathtt{LIFT},\mathtt{REORIENT},\\
&\mathtt{TRANSFER},\mathtt{PREINSERT\_DWELL},\mathtt{INSERT},\mathtt{INSERT\_DWELL},\mathtt{RELEASE},\mathtt{RETREAT},\mathtt{SETTLE}\},\\
\mathcal M={}&\{\mathtt{LIFT},\mathtt{REORIENT},\mathtt{TRANSFER},\mathtt{PREINSERT\_DWELL},\mathtt{INSERT},\mathtt{INSERT\_DWELL},\mathtt{RELEASE},\mathtt{RETREAT}\}.
\end{aligned}
\end{equation*}
The phases in $\mathcal K\setminus\mathcal M$ retain their nominal durations. These fixed phases contribute $6.3$~s and the phases in $\mathcal M$ contribute $7.8/r$~s, so the cycle lasts $14.1$~s at $r=1$ and $10.2$~s at $r=2$. Task geometry, waypoints, physical parameters, and task tolerances remain fixed. \Cref{fig:system} summarizes the task and phase-duration scaling.

\subsubsection{Expert, Demonstrations, and Learned Policies}\label{ssec:ratecalib}
The expert $\pi_E$ generates the demonstrations and provides the reference success rates. Its waypoint sequence implements the approach, acquisition, lift, reorientation, transfer, pose before insertion, insertion, release, and retreat geometry. These waypoints and all controller parameters remain fixed across execution speeds; only the phase schedule changes through \Cref{eq:rate}. The parcel moves only through simulated contact forces. We selected $\mathcal R_D=[0.5,2]$ based on a calibration of the expert performed prior to learner evaluation. Across 64 episodes per candidate speedup factor with $1$~cm planar jitter in parcel position, the expert completes $63/64$ episodes at $r=1$, $64/64$ at $r=1.5$, $58/64$ at $r=2$, $19/64$ at $r=2.5$, and $1/64$ at $r=3$ (\Cref{app:calibration}). The selected boundary therefore does not depend on learner outcomes.

We collect 300 expert episodes with $r\sim\mathcal U[0.5,2]$. Independent planar parcel-position offsets are sampled uniformly from $[-10,10]$~mm on each axis, and robot joint-position offsets are sampled uniformly from $[-0.05,0.05]$~rad. Robot and parcel velocities are zero at reset; mass and friction are fixed. The 297 successful episodes form the common demonstration set. We train DAgger \cite{ross2011dagger}, Diffusion Policy \cite{chi2023diffusion}, and ACT \cite{zhao2023act} on these observations and actions. We omit vision-language-action baselines such as $\pi_0$ \cite{black2024pi0} and GR00T~N1 \cite{nvidia2025gr00t} because their released models require camera observations, whereas ParcelStow supplies state observations.

We train three ACT policies with common demonstrations, architecture, optimizer, batch construction, loss, temporal ensembling, and 2000-epoch budget. They differ only in pseudorandom parameter initialization; we denote them ACT-A, ACT-B, and ACT-C. Each uses a chunk length of 100 and a transformer with 4 encoder and 7 decoder layers. \Cref{app:act} gives the complete training configuration.

\subsection{Evaluation Protocol}\label{ssec:evalproto}
We set $\mathcal R_E=\{0.5,1,1.5,2,2.25,2.5,3\}$. Each policy--speed cell contains 100 episodes. At each value of $r$, all policies use the same ordered initial-condition draws from the reset distribution used for demonstration collection. A shared evaluation seed reproduces these parcel and robot joint offsets, yielding 100 matched expert--learner episode pairs at every common speed. Observation corruption is disabled.

For each policy and speedup factor, $\widehat p_\pi(r)$ is the number of successful episodes divided by 100. We report a $95\%$ Wilson confidence interval for each estimate. For differences between policies, we resample the 100 matched episode pairs with replacement, recompute the signed difference for each of $20{,}000$ paired bootstrap samples, and report the central $95\%$ interval. We also report each policy's change from $\widehat p_\pi(1)$.

Alongside success of the overall task, we record stage completion, terminal failure reason, hand--parcel relative motion under \Cref{eq:relmotion_diag}, handoff outcome under \Cref{eq:relative_handoff_command}, realized contact margins, and the maximum arm joint velocity utilization
\begin{equation}
U_{\mathrm{arm}}=\max_{t\ge t_{\mathrm{lift}}}\ \max_{j\in\mathcal J_{\mathrm{arm}}}\ |\dot q_j(t)|\,/\,\dot q_{j,\max}.
\label{eq:armutil}
\end{equation}
$t_{\mathrm{lift}}$ denotes the first control step of the $\mathtt{LIFT}$ phase, $\mathcal J_{\mathrm{arm}}$ is the set of arm joints, and $\dot q_{j,\max}$ is the velocity limit of joint $j$.
Tests of simulator integrity exercise the dynamics of the free parcel, receptacle contact model, success predicates, and phase-duration law independently of any learner.

\subsection{Failure Diagnostics}\label{ssec:diagnostics}
The expert--learner success rate difference identifies performance separation but does not locate the first failure or establish a contribution from parcel retention. Stage outcomes locate the first failed predicate, and the relative-motion handoff tests retention under shared hand motion after acquisition. Hand--parcel relative motion and the force closure margin characterize the realized grasp; neither defines success of the overall task. The recorded-rollout diagnostics read simulator state, whereas the handoff experiment replaces downstream hand commands. Neither changes the success predicates or selected speedup factors.

\subsubsection{Stage Outcomes}\label{ssec:stageoutcomes}
The ordered success predicates of \Cref{ssec:embtsk} identify the last task stage completed by each episode. We report acquisition, lift, reorientation, pose before insertion, insertion, release, and settling outcomes, together with the terminal failure reason assigned by the simulator state machine. The state machine classifies an episode as an insertion jam when the parcel reaches the pose before insertion but does not satisfy the insertion predicate and either maintains parcel--receptacle contact above $2$~N for at least $0.2$~s or exceeds the hand--parcel relative-motion limit only after receptacle contact. These outcomes locate the first failed predicate but do not identify its physical cause.

\subsubsection{Relative-Motion Handoff}\label{ssec:relmothndff}
The relative-motion handoff replaces each policy's commands after acquisition while preserving its realized hand shape and hand pose at the handoff instant. Each policy executes the acquisition phases on its own, after which a shared controller drives the hand along the expert's subsequent motion, expressed relative to the expert's hand pose at handoff. Let $T^{W,E}_H(s)\in\mathrm{SE}(3)$ denote the expert hand pose in the world frame $W$ at time $s$ after handoff. The expert's relative hand motion is
\begin{equation}
\Delta T_H^E(s)=\left(T^{W,E}_{H}(0)\right)^{-1}T^{W,E}_{H}(s),
\label{eq:relative_handoff}
\end{equation}
and the shared controller commands the evaluated policy's hand to
\begin{equation}
T^{W,d}_{H,\pi}(s)=T^W_{H,\pi}(0)\,\Delta T_H^E(s),
\label{eq:relative_handoff_command}
\end{equation}
where $T^W_{H,\pi}(0)$ is the policy's own realized hand pose at handoff and the superscript $d$ marks the commanded target. Anchoring at $T^W_{H,\pi}(0)$ preserves the policy's realized hand pose at acquisition while replacing its downstream hand motion with the expert's relative hand motion.

The primary endpoint is parcel retention through reorientation and transfer, evaluated until insertion begins or the parcel first contacts the receptacle. Retention under \Cref{eq:relative_handoff_command} tests each acquired grasp under the expert's relative hand motion. The secondary endpoint continues the same relative-motion controller through insertion, release, and settling. Differences at this endpoint can arise from the hand and parcel state at handoff or from their subsequent interaction with the receptacle; the experiment does not separate these effects. Because the handoff replaces the learned downstream controller, it also cannot attribute a residual difference to that controller.

\subsubsection{Hand--Parcel Relative Motion}\label{ssec:handobjmotion}
Retention is a binary outcome; hand--parcel relative motion quantifies continuous changes in the acquired parcel pose. Denoting $T^W_{H,t}$ and $T^W_{O,t}$ as the hand and object poses in the world frame, the object pose in the hand frame is $T^H_{O,t}=(T^W_{H,t})^{-1}T^W_{O,t}$. We let $t_G$ denote the first stable acquired state after closure and write the hand--parcel relative motion as
\begin{equation}
\Xi_{HO}(t)=\left(T^H_{O,t_G}\right)^{-1}T^H_{O,t},
\label{eq:relmotion}
\end{equation}
where $R_{\Xi}(t)$ is the rotation block of $\Xi_{HO}(t)$. We report the peak hand--parcel relative translation and rotation
\begin{equation}
d_p=\max_{t\ge t_G}\left\|\operatorname{trans}\!\left(\Xi_{HO}(t)\right)\right\|_2,
\qquad
d_R=\max_{t\ge t_G}\left\|\log\!\left(R_{\Xi}(t)\right)^\vee\right\|_2 .
\label{eq:relmotion_diag}
\end{equation}
Both quantities characterize changes in the acquired parcel pose, and neither defines task failure.

\subsubsection{Force Closure at Acquisition}\label{ssec:forceclosure}
Force closure characterizes the ability of an episode's realized fingertip contact set to span the wrench space under the nominal friction coefficient. At acquisition, at the end of reorientation, and at insertion start, the recorded contacts supply points, normals, and forces, and we score each set with the Ferrari--Canny margin $\FC$ \cite{ferrari1992planning}. We write $\FC>0$ for force closure and $\FC\le 0$ otherwise. The margin describes the contacts realized in simulation, not a planned or idealized grasp. We use the margin in one direction only. The sign test $\FC\le 0$ has a physically defined threshold and zero fitted parameters, and we evaluate it as a candidate necessary condition for downstream success. We evaluate the continuous magnitude separately in \Cref{app:oos} through held-out ranking and calibration across execution speeds.

\section{Results and Discussion}\label{sec:results}
\subsection{Expert and ACT-A Differ Across the Demonstrated Speed Range}\label{ssec:envelopes}
\begin{figure}[t]
\centering
\includegraphics[width=0.86\textwidth]{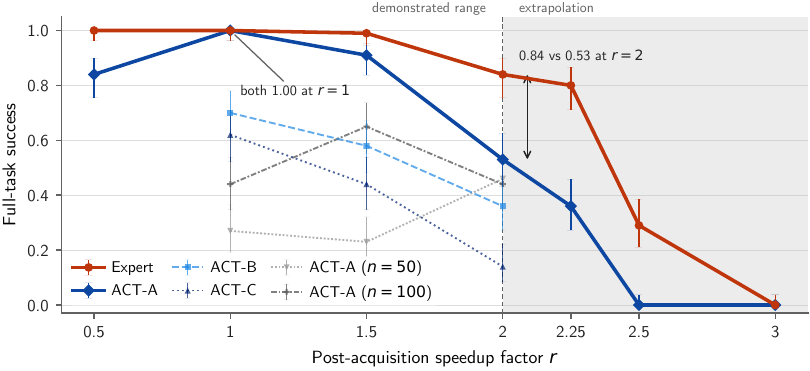}
\caption{\textbf{Task Success Across Execution Speeds.} Task success against the speedup factor for the phases after acquisition $r$ with Wilson $95\%$ intervals, 100 episodes per point. The expert and ACT-A coincide at $r=1$ and differ across the demonstrated speed range $r\le2$. ACT-B and ACT-C show larger decreases in success over $r\in[1,2]$ than the expert while starting below expert success at $r=1$. Gray curves show ACT-A's configuration retrained on 50 and 100 demonstrations, where nominal success increases with dataset size while success at $r=2$ changes little. The shaded band marks extrapolation beyond the demonstrated range, $r>2$.}
\label{fig:capability}
\end{figure}

\Cref{tab:rates} and \Cref{fig:capability} report success of the overall task across the demonstrated speed range. The expert and ACT-A each complete 100 of 100 episodes at $r=1$. At the interior speed $r=1.5$, the expert completes 99 episodes and ACT-A completes 91. At the maximum demonstrated speed, $r=2$, the counts fall to 84 and 53, with Wilson intervals of $[0.76,0.90]$ and $[0.43,0.63]$. From $r=1$ to $r=2$, expert success decreases by 16 percentage points and ACT-A success decreases by 47. Their success rates therefore differ by 31 percentage points at $r=2$. Over the 100 matched pairs of initial conditions, the paired bootstrap places a $95\%$ interval of $[0.18,0.44]$ on this difference.

\begin{table}[t]
\centering
\small
\setlength{\tabcolsep}{7pt}
\caption{{Task success over the demonstrated speed range.} Each cell reports the success rate over 100 episodes. The final column reports the decrease from $r=1$ to $r=2$. Mint marks the expert and ACT-A cells with equal success at $r=1$, and salmon highlights ACT-A at $r=2$, the primary expert--learner comparison. ACT-B, ACT-C, Diffusion Policy, and DAgger begin below expert success at $r=1$, so their expert--learner differences at higher speeds include the difference already present at nominal speed.}
\label{tab:rates}
\begin{tabular}{@{}lcccc@{}}
\toprule
Policy & $r=1$ & $r=1.5$ & $r=2$ & Decrease \\
\midrule
Scripted expert & \cellcolor{oursmint}1.00 & 0.99 & 0.84 & 0.16 \\
ACT-A & \cellcolor{oursmint}1.00 & 0.91 & \cellcolor{losssalmon}0.53 & 0.47 \\
ACT-B & 0.70 & 0.58 & 0.36 & 0.34 \\
ACT-C & 0.62 & 0.44 & 0.14 & 0.48 \\
\midrule
Diffusion Policy & 0.68 & 0.73 & 0.54 & 0.14 \\
DAgger & 0.03 & 0.01 & 0.03 & 0.00 \\
\bottomrule
\end{tabular}
\end{table}

\begin{figure}[t]
\centering
\begin{tabular}{@{}cc@{}}
\includegraphics[width=0.59\textwidth]{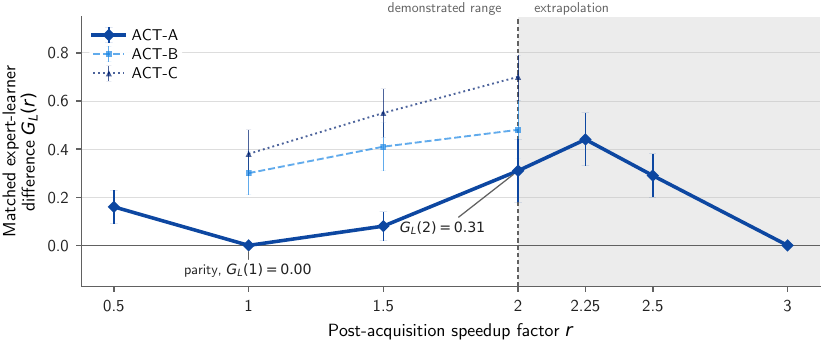} &
\includegraphics[width=0.35\textwidth]{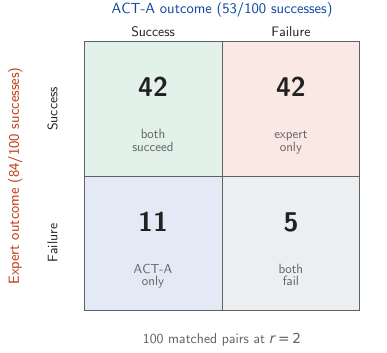} \\
\small (a) Matched success rate difference & \small (b) Matched outcomes at $r=2$
\end{tabular}
\caption{\textbf{Matched Expert--Learner Comparison at Curve and Episode Levels.} (a) $G_L(r)=\widehat p_{\pi_E}(r)-\widehat p_{\pi_L}(r)$ with paired bootstrap $95\%$ intervals over matched pairs of initial conditions. ACT-A is shown over the full evaluation grid and ACT-B/C over $r\in\{1,1.5,2\}$. ACT-A has equal success with the expert at $r=1$, but the difference reaches $0.31$ at $r=2$ with interval $[0.18,0.44]$; its nonzero value at $r=0.5$ shows that the equality at $r=1$ does not persist across the demonstrated speed range. (b) The $2\times2$ matrix of matched outcomes for the expert and ACT-A at $r=2$: both succeed on 42 episodes, the expert alone succeeds on 42, ACT-A alone succeeds on 11, and both fail on 5, giving marginal success counts of 84 and 53. ACT-A succeeds on 11 episodes in which the expert fails, so ACT-A's failures are not a superset of the expert's failures.}
\label{fig:matchedviews}
\end{figure}

ACT-A coincides with the expert at $r=1$ but differs at both slower and faster demonstrated speeds, including a difference of $0.16$ at $r=0.5$ and $0.31$ at $r=2$. At $r=2$, the episode-level comparison shows reversals in both directions: 42 matched initial conditions succeed only for the expert, while 11 succeed only for ACT-A.

ACT-B and ACT-C start below expert success at $r=1$, with success rates of 0.70 and 0.62. From $r=1$ to $r=2$ their success decreases by 0.34 and 0.48, compared with 0.16 for the expert. Thus, all three ACT policies show larger decreases in success than the expert over this speed change, although their nominal success ranges from 0.62 to 1.00. Diffusion Policy records 0.68, 0.73, and 0.54 over the same three speeds, so its success is not monotone in $r$. DAgger records 0.03, 0.01, and 0.03; its proximity to the success-rate floor precludes a meaningful comparison of speed dependence.

Beyond the demonstrated speed range, the expert--ACT-A success difference is 44 percentage points at $r=2.25$ and 29 percentage points at $r=2.5$. At $r=3$, neither policy completes an episode, so the difference is zero at the floor of the task. As expert success decreases beyond $r=2$, minimum orientation error before insertion increases from $2.2^\circ$ at $r=1$ to $8.9^\circ$ at $r=2.25$ and $10.3^\circ$ at $r=2.5$, crossing the $10^\circ$ tolerance over the interval where expert success falls from 0.80 to 0.29. Over the same interval, median insertion depth falls from $56.3$ to $49.1$~mm and median peak receptacle contact force rises from $3.1$ to $8.1$~N. Arm joint velocity utilization remains below $0.17$ and median peak hand--parcel translation remains below $9$~mm through $r=2.5$. The decline in expert success coincides with greater orientation error before insertion, lower insertion depth, and higher receptacle contact force; arm joint velocity utilization remains far below its limit, and parcel loss is not observed. \Cref{app:ceiling} reports the full grid.

To examine sensitivity to the number of training demonstrations, we retrain ACT-A's configuration and parameter initialization on subsets of 50 and 100 demonstrations stratified by speedup factor. Nominal success is 0.27, 0.44, and 1.00 for 50, 100, and 297 demonstrations, respectively. At $r=2$, success is 0.46, 0.44, and 0.53, compared with 0.84 for the expert. Across the three training-set sizes, more demonstrations increase nominal success but do not produce a corresponding increase at $r=2$. The experiment does not test training sets larger than 297 demonstrations. \Cref{app:scaling} gives the subset protocol and individual cells. ACT-A matches the expert at $r=1$ yet differs by 31 percentage points at the maximum demonstrated speed, $r=2$; ACT-B and ACT-C also show larger decreases in success than the expert from $r=1$ to $r=2$, although neither matches expert success at nominal speed.

\subsection{Most ACT-A Failures Occur During Insertion}\label{ssec:localization}
\begin{figure}[t]
\centering
\includegraphics[width=0.94\textwidth]{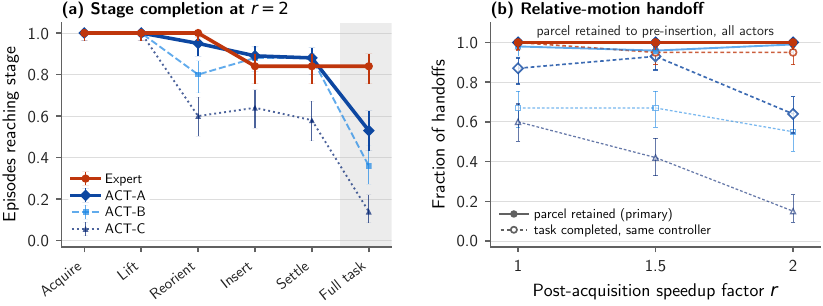}
\caption{\textbf{Stage Completion and Relative-Motion Handoff.} (a) Fraction of episodes reaching each latched stage at $r=2$ for the expert and the three ACT policies. The expert and ACT-A coincide through acquisition and lift; their difference first appears at reorientation and is largest after insertion. (b) Solid curves report parcel retention through reorientation and transfer in free space under the expert's relative hand motion from \Cref{eq:relative_handoff_command}. Dashed curves report success of the overall task after the same relative-motion law continues through insertion, release, and settling.}
\label{fig:attribution}
\end{figure}

\begin{figure}[t]
\centering
\begin{subfigure}[t]{0.47\textwidth}
\centering
\includegraphics[width=\linewidth]{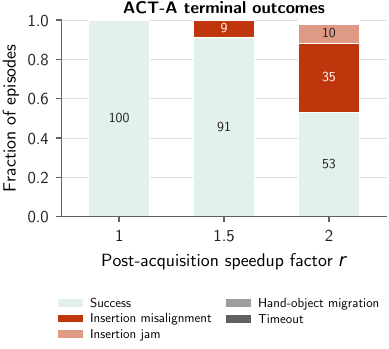}
\caption{\textbf{Terminal Outcomes Across Demonstrated Speeds.} Outcome composition for ACT-A at $r\in\{1,1.5,2\}$.}
\label{fig:failuremodes}
\end{subfigure}\hfill
\begin{subfigure}[t]{0.49\textwidth}
\centering
\includegraphics[width=\linewidth]{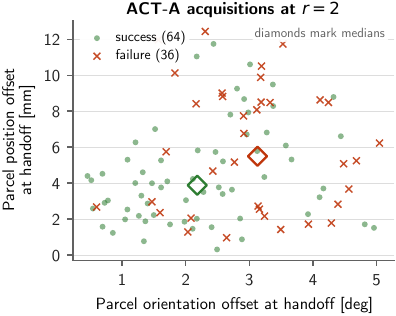}
\caption{\textbf{State at Handoff Offset and Downstream Outcome at $r=2$.} ACT-A parcel-pose offsets relative to the matched expert at handoff.}
\label{fig:handoffstate}
\end{subfigure}
\caption{\textbf{ACT-A Failure Localization and Diagnostics of the State at Handoff.} (a) Normalized terminal outcomes show 100 successes at $r=1$, nine insertion misalignments at $r=1.5$, and at $r=2$ a composition of 53 successes, 35 insertion misalignments, 10 insertion jams, one failure classified as excessive parcel slip, and one timeout. The increasing failure rate is therefore dominated by insertion failures. (b) At $r=2$, failed handoffs have larger median rotational and translational offsets: $3.1^\circ$ and $5.5$~mm for failures, compared with $2.2^\circ$ and $3.9$~mm for successes. The groups overlap substantially, preventing the handoff-state offsets from discriminating success and failure reliably.}
\label{fig:failurehandoff}
\end{figure}

Arm joint velocity utilization remains far below its limit at $r=2$. Median $U_{\mathrm{arm}}$ is $0.108$ for the expert and $0.107$ for ACT-A, with 90th percentiles of $0.111$ and $0.117$ and median peak hand linear speeds of $0.371$ and $0.378$~m/s. These records contain arm joint velocities but not joint torques, acceleration-dependent tracking error, hand-joint limits, or contact dynamics. \Cref{fig:attribution}a localizes the expert--ACT-A difference by task stage, while \Cref{fig:failuremodes} shows how ACT-A's terminal outcomes change across execution speeds. Under ACT-A at $r=2$, acquisition and lift succeed in all 100 episodes, reorientation succeeds in 95, and 94 episodes reach pose before insertion. Of the 89 episodes satisfying the latched insertion predicate, 53 also satisfy the terminal position and orientation tolerances after release and settling. The 47 failures comprise 35 insertion misalignments, 10 insertion jams, one failure classified as excessive parcel slip, and one timeout. The expert reaches pose before insertion in all 100 episodes; 16 subsequently fail because of insertion jams.

The expert--ACT-A difference first appears at reorientation, where ACT-A loses 5 episodes and the expert loses none. At pose before insertion, the difference is 6 percentage points: 94 ACT-A episodes and all 100 expert episodes reach the stage. Task success differs by 31 percentage points, so 25 points of the final difference arise after pose before insertion. ACT-A records 35 insertion misalignments and 10 insertion jams, while the expert records 16 insertion jams.

\subsection{Shared Motion Preserves Free Space Parcel Retention}\label{ssec:handoffresults}
\Cref{fig:attribution}b reports the relative-motion handoff results described in \Cref{ssec:relmothndff}. Across $r\in\{1,1.5,2\}$ the expert, ACT-A, and ACT-C each acquire the parcel in all 300 episodes and retain it in all 300 under the expert's relative hand motion. ACT-B acquires 100, 98, and 100 episodes and retains 98, 94, and 99 of them. Median hand--parcel relative translation stays below $5.8$~mm and median hand--parcel relative rotation below $5.4^\circ$ across policies and execution speeds. ACT-A therefore retains the parcel through reorientation and transfer in free space under this imposed motion; parcel loss during this interval does not account for its expert--learner success-rate difference.

Continuing the expert's relative motion through insertion, release, and settling at $r=2$ yields 95 successes after expert acquisition and 64, 55, and 15 after ACT-A, ACT-B, and ACT-C acquisition. For ACT-A, 64 handoff episodes complete the task, compared with 53 end-to-end episodes under ACT-A's own downstream commands. Expert and ACT-A acquisitions use the same relative-motion law, anchored at each policy's realized hand pose. Their parcel poses and other physical states at handoff can still differ. The 64-versus-95 result therefore shows that replacing ACT-A's downstream commands does not remove the success difference, but it does not separate the state at handoff from subsequent receptacle contact. In $65\%$ of ACT-C episodes at $r=2$, the parcel contacts the receptacle before the nominal pose before insertion endpoint, and the shared motion carries the pre-insertion pose offset into receptacle contact. This ACT-C result does not explain ACT-A's handoff failures, and the replacement does not evaluate any learner's downstream controller.

Failed ACT-A handoffs have larger median pose offsets than successful ones, but the distributions in \Cref{fig:handoffstate} overlap substantially. The overlap prevents attribution of the failures to handoff-state error alone; tracking error, hand actuation, and subsequent receptacle contact also remain unresolved.

Acquisition and lift complete in every ACT-A episode, arm joint speeds remain far below their limits, and every ACT-A acquisition retains the parcel through reorientation and transfer in free space under the expert's relative hand motion. Thirty-five of ACT-A's 47 failures are insertion misalignments. The failures are therefore concentrated during insertion, but the present diagnostics do not distinguish among handoff-state error, tracking error, hand actuation, and receptacle contact.

\subsection{No Acquisition Without Force Closure Succeeds}\label{ssec:certificate}
Across six policies and every evaluated speedup factor, none of the 414 acquired episodes with $\FC\le 0$ at acquisition completes the task, placing a $95\%$ Wilson upper bound of $0.009$ on success without force closure. All 700 expert acquisitions satisfy force closure. The same one-sided result holds for the two checkpoints trained with different numbers of demonstrations: the additional 79 acquisitions without force closure all fail. Acquisitions without force closure appear in $8\%$ of ACT-A episodes, $21\%$ of ACT-B episodes, and $34\%$ of ACT-C episodes.

The a priori force closure test uses the physical threshold $\FC=0$ and contains no fitted parameters. Separately, fitting a Youden threshold on speeds within the demonstrated range yields a value within $10^{-3}$ of zero in every reported protocol. The discriminative performance of the continuous margin varies across test populations: held-out AUROC is $0.62$ at acquisition and $0.72$ at insertion start, ranges from $0.70$ to $0.83$ under leave-one-policy-out evaluation, and is $0.70$ and $0.83$ on ACT-B and ACT-C. A single logistic model shared across execution speeds has a held-out expected calibration error of $0.31$. Thus, absence of force closure at acquisition is associated with failure in these evaluations, while the continuous margin does not rank success reliably or yield well-calibrated success probabilities across speeds and policies. \Cref{app:oos} reports the full held-out analysis.

\subsection{Interpretive Scope and Open Mechanisms}\label{sec:discussion}
One untested mechanism is ACT's temporally extended action prediction. ACT predicts action chunks with temporal ensembling, while the expert computes a new command at every control step \cite{zhao2023act}. Our experiments do not vary chunk horizon, temporal ensembling, or replanning frequency and therefore cannot attribute the expert--ACT-A success-rate difference to action chunking. The evaluation varies execution speed while holding task geometry, physical parameters, and success criteria fixed for one contact-rich task with state observations and fixed-friction simulation. The results therefore do not establish generalization of this speed sensitivity to other architectures, tasks, observation modalities, or physical hardware. Within the tested setting, force closure is necessary for task success but not sufficient for it.

\section{Conclusion}\label{sec:conclusion}
We formulated an expert--learner task success comparison across execution speeds and instantiated it in ParcelStow. ACT-A and the scripted expert both achieve 100\% task success at nominal speed, yet their success rates differ by 31 percentage points at the maximum demonstrated speed. From $r=1$ to $r=2$, success also decreases more for two additional ACT policies than for the expert, although those policies begin below expert success. Stage outcomes place most ACT-A failures during insertion, while the relative-motion handoff shows that its acquisitions retain the parcel during free-space transport under shared hand motion. No evaluated acquisition without force closure succeeds, but the continuous force closure margin does not consistently rank task success. These results show that equal nominal expert--learner success can conceal different sensitivity to execution speed in contact-rich manipulation. Experiments with other architectures, tasks, sensing modalities, and physical hardware are needed to assess generality.

\bibliographystyle{ieeetr}
\bibliography{references}

\clearpage
\appendix
\section*{Appendix}
\section{Physical Task Specification}\label{app:task}
\Cref{tab:app_task} lists the ParcelStow quantities specified before learner evaluation. We determined these values from kinematic probes and the expert-only calibration in \Cref{app:calibration}; no learner outcome informed them. The dynamics of the free parcel and the regression tests of \Cref{ssec:embtsk} apply unchanged, and the success predicates contain no wrench-space quantity.

\begin{table}[htb]
\centering
\small
\caption{ParcelStow Quantities Specified Before Learner Evaluation.}
\label{tab:app_task}
\begin{tabular}{@{}ll@{}}
\toprule
Quantity & Value \\
\midrule
Parcel extents and mass & $80\times55\times40$~mm, $0.12$~kg \\
Nominal friction & $\mu=0.5$ static and dynamic \\
Start pose & $(0.35,\,0,\,0.721)$~m, yaw $45^\circ$ \\
Reorientation & $90^\circ$ about the parcel width axis \\
Transport distance & $0.334$~m \\
Tight-axis insertion clearance & $10$~mm per side \\
Final orientation tolerance & $10^\circ$ \\
Policy and control frequency & $50$~Hz \\
Training speedup distribution & $r\sim\mathcal U[0.5,2.0]$ \\
Evaluation speedup grid & $\{0.5,1,1.5,2,2.25,2.5,3\}$ \\
Fixed-phase duration & $6.3$~s \\
Duration of phases in $\mathcal M$ & $7.8/r$~s \\
\bottomrule
\end{tabular}
\end{table}

\Cref{tab:app_predicates} states the success predicates evaluated on simulator state.

\begin{table}[htb]
\centering
\small
\caption{Success Predicates of the Overall Task. Insertion and settling latch as stage events, and success of the overall task additionally requires the terminal position and orientation criterion.}
\label{tab:app_predicates}
\begin{tabular}{@{}p{0.20\textwidth}p{0.74\textwidth}@{}}
\toprule
Predicate & Condition \\
\midrule
Acquisition & Parcel center rises by at least $20$~mm while the thumb distal phalanx and at least one further distal phalanx each exert more than $1$~N of contact force on the parcel. \\
Lift clearance & Parcel center rises at least $60$~mm above the table. \\
Reorientation & Parcel orientation error falls below $15^\circ$. \\
Pre-insertion & Parcel orientation error falls below $15^\circ$ before insertion. \\
Insertion & Parcel center passes at least $50$~mm beyond the receptacle entrance plane while remaining inside the receptacle cross section. \\
Release & Contact force between the distal phalanges and parcel falls below $0.5$~N for at least $0.1$~s. \\
Settling & Released parcel remains inserted for $0.4$~s with linear speed below $0.02$~m/s, angular speed below $0.2$~rad/s, and final orientation error below $10^\circ$. \\
\bottomrule
\end{tabular}
\end{table}

\section{Observation, Action, and Control Details}\label{app:control}
The observation $o_t\in\mathcal O\subset\mathbb R^{147}$ contains full-body joint position and velocity (51 values each), the previous action $u_{t-1}$ (16), parcel pose in the base frame (7), right-hand fingertip positions (15), right-hand fingertip contact forces (5), task phase (1), and speedup factor $r$ (1). The 51 joint values cover every actuated and passively coupled degree of freedom in the G1 and L6 articulation; the fingertip terms cover five distal phalanges. At episode reset, the environment resets the action manager before constructing the first observation, so the initial previous-action term is the action manager's reset state. The action $u_t\in\mathcal U\subset\mathbb R^{16}$ contains normalized joint position offsets for the waist (3), right arm and wrist (7), and six actuated right-hand joints (thumb roll and pitch and four finger MCP-pitch joints). The left arm, remaining hand joints, and passive coupling joints appear in the observation without corresponding action terms.

Implicit PD joint drives track $q_{\mathrm{target}}=q_{\mathrm{default}}+0.5\,u_t$ with gains of $K_p{=}5000$, $K_d{=}5$ for the waist; $K_p{=}300$, $K_d{=}10$ for the arm; and $K_p{=}10$, $K_d{=}0.2$ for the hand. The control frequency is $50$~Hz. Simulation runs at $200$~Hz ($0.005$~s) with $4\times$ decimation, a fixed pelvis, 16 position solver iterations, and 8 velocity solver iterations.

\paragraph{Scripted Expert.}
The expert's waypoint sequence in task space is converted offline to waist and right arm joint-space knots by damped least-squares differential inverse kinematics with damping $\lambda{=}0.05$. The solver uses a null-space mid-range bias and a $0.15$-rad step clamp and iterates until the waypoint solution reaches $4$~mm position and $2^\circ$ orientation tolerance. At runtime, cosine interpolation between adjacent knots generates the phase reference, and a small integral correction ($k_i{=}0.08$, clamped to $\pm0.35$) compensates residual tracking error before the command is emitted through the common joint position interface. The expert generates a command at every $50$~Hz policy step rather than committing to a multi-step action chunk. The geometric knots, inverse kinematics tolerances, and controller parameters remain unchanged across speedup factors; $r$ changes the scheduled durations of the phases after acquisition according to \Cref{eq:rate}. Consequently, increasing $r$ reduces the time available to traverse the same geometric reference and changes the resulting velocities, accelerations, tracking demands, and contact transients without changing the expert's task geometry.

\section{ACT Architecture and Training}\label{app:act}
Each ACT instance uses a chunk length of 100, a transformer encoder of 4 layers and a decoder of 7 layers, 8 attention heads, a model dimension of 512, a feedforward dimension of 3200, a 32-dimensional latent variable, and dropout 0.1. Training uses AdamW at learning rate $10^{-5}$ with weight decay $10^{-4}$, batch size 8, and a KL weight of 10, for 2000 epochs over the shared 297-episode demonstration set. Evaluation applies temporal ensembling over the predicted chunks. The observation vector includes the speedup factor $r$ alongside the proprioceptive state, making $r$ available to the policy at every control step. \Cref{tab:app_act} reports the final training loss of each instance.

\begin{table}[htb]
\centering
\small
\caption{ACT training loss at 2000 epochs, by pseudorandom parameter initialization.}
\label{tab:app_act}
\begin{tabular}{@{}lccc@{}}
\toprule
 & ACT-A & ACT-B & ACT-C \\
\midrule
\rowcolor{oursmint} Final training loss & 0.0388 & 0.0403 & 0.0405 \\
\bottomrule
\end{tabular}
\end{table}

\section{Speed Calibration Using the Expert}\label{app:calibration}
\begin{figure}[htb]
\centering
\includegraphics[width=0.92\textwidth]{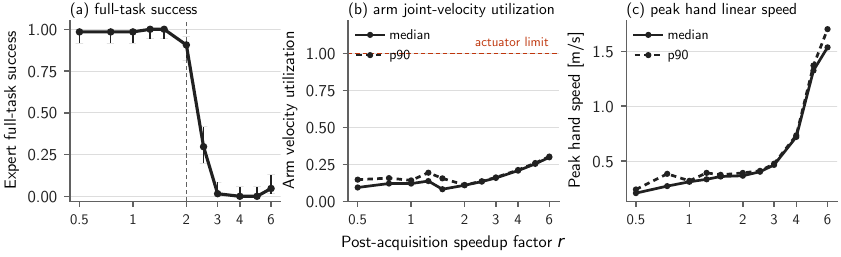}
\caption{\textbf{Speed Calibration Using the Expert Over the Candidate Grid.} (a) Task success with Wilson $95\%$ intervals. (b) Maximum arm joint velocity utilization, median and 90th percentile, against the actuator limit. (c) Peak hand linear speed, median and 90th percentile. The dashed line marks the maximum of the demonstrated speed range, $r=2$, chosen without observing any learner outcome.}
\label{fig:app_sweep}
\end{figure}

We selected the demonstrated and evaluation speed ranges using expert-only calibration. \Cref{tab:app_calibration} reports 64 expert episodes at each candidate speedup factor under $1$~cm planar jitter in the start pose, and \Cref{fig:app_sweep} plots the same calibration with arm joint velocity utilization and peak hand speed. Expert success remains at or above $58/64$ through $r=2$, then falls to $19/64$ at $r=2.5$ and $1/64$ at $r=3$. We selected $[0.5,2]$ as the support of the demonstrated speed distribution and treat larger values as extrapolation beyond that support. Arm joint velocity utilization remains below $0.3$ across the candidate grid; velocity saturation therefore does not set the upper demonstrated-speed boundary. The three successes at $r=6$ are reported for completeness but lie outside the demonstrated and evaluation speed ranges analyzed in the paper.

\begin{table}[htb]
\centering
\small
\setlength{\tabcolsep}{4pt}
\caption{Speed calibration using the expert, with 64 episodes per speedup factor, performed prior to learner evaluation. Mint denotes values with uniform success, while salmon denotes values with complete failure.}
\label{tab:app_calibration}
\begin{tabular}{@{}lccccccccccc@{}}
\toprule
$r$ & 0.5 & 0.75 & 1.0 & 1.25 & 1.5 & 2.0 & 2.5 & 3.0 & 4.0 & 5.0 & 6.0 \\
\midrule
Success & 63/64 & 63/64 & 63/64 & \cellcolor{oursmint}64/64 & \cellcolor{oursmint}64/64 & 58/64 & 19/64 & 1/64 & \cellcolor{losssalmon}0/64 & \cellcolor{losssalmon}0/64 & 3/64 \\
\bottomrule
\end{tabular}
\end{table}

\begin{figure}[htb]
\centering
\includegraphics[width=\textwidth]{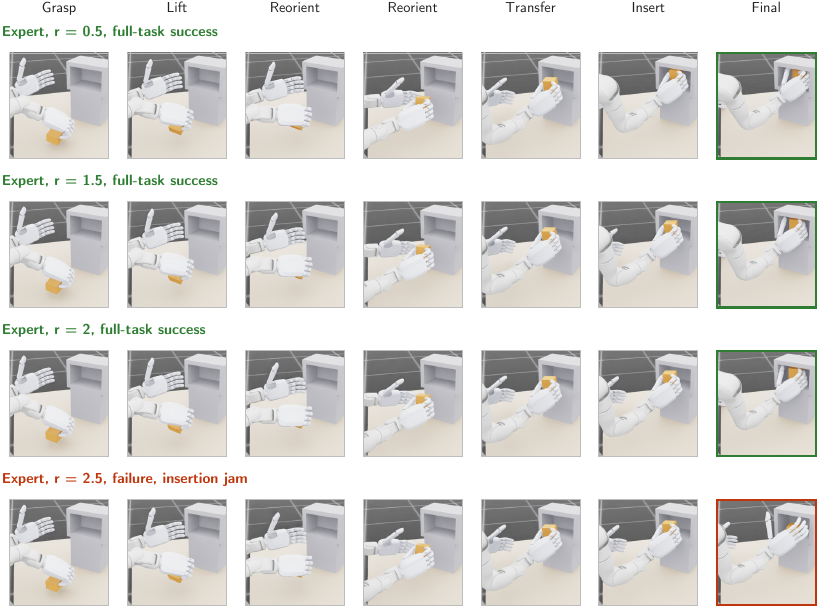}
\caption{\textbf{Expert Execution Across the Calibrated Speed Range.} Task geometry remains fixed while the expert traverses the same manipulation under contracting time budgets.}
\label{fig:app_rates}
\end{figure}

\section{Expert Diagnostics at Higher Execution Speeds}\label{app:ceiling}
\Cref{tab:app_ceiling} reports medians over the 100-episode expert evaluation cells at each speedup factor. Minimum orientation error before insertion crosses the $10^\circ$ tolerance between $r=2.25$ and $r=2.5$, the interval where expert success decreases from 0.80 to 0.29, while arm joint velocity utilization remains below 0.13 and peak hand--parcel translation remains below $9$~mm. The near-unity full-body joint velocity utilization at every value of $r$, including $r=0.5$, is due primarily to finger motion during the fixed-duration grasp-close phase. Because this utilization does not increase with $r$, it cannot explain the decline in expert success at higher execution speeds.

\begin{table}[htb]
\centering
\small
\setlength{\tabcolsep}{6pt}
\caption{Expert success and diagnostic quantities over the evaluation grid, reported as medians over 100 episodes per speedup factor. Mint marks uniform success and salmon complete failure. Bold indicates speeds at which the minimum orientation error before insertion exceeds the $10^\circ$ tolerance.}
\label{tab:app_ceiling}
\begin{tabular}{@{}lccccccc@{}}
\toprule
$r$ & 0.5 & 1 & 1.5 & 2 & 2.25 & 2.5 & 3 \\
\midrule
Success & \cellcolor{oursmint}1.00 & \cellcolor{oursmint}1.00 & 0.99 & 0.84 & 0.80 & 0.29 & \cellcolor{losssalmon}0.00 \\
Min.\ Pre-Insertion Orientation Error [$^\circ$] & 2.8 & 2.2 & 4.7 & 7.6 & 8.9 & \cellcolor{losssalmon}\textbf{10.3} & \cellcolor{losssalmon}\textbf{10.8} \\
Insertion Depth [mm] & 63.7 & 56.3 & 52.5 & 50.0 & 49.5 & 49.1 & 30.4 \\
Peak Receptacle Contact Force [N] & 2.7 & 3.1 & 4.7 & 5.7 & 7.0 & 8.1 & 13.6 \\
Arm Joint Velocity Utilization & 0.10 & 0.12 & 0.08 & 0.11 & 0.12 & 0.13 & 0.16 \\
Peak Hand--Parcel Translation [mm] & 5.8 & 8.2 & 8.2 & 8.1 & 8.2 & 8.5 & 20.8 \\
\bottomrule
\end{tabular}
\end{table}

\section{Demonstration Scaling}\label{app:scaling}
For each training-set size, we retrain ACT-A with the same architecture, optimizer, batch construction, loss, temporal ensembling, 2000-epoch budget, and pseudorandom parameter initialization. To construct a subset of size $n$, we sort the 297 demonstrations by speedup factor and select evenly spaced ranks, so every subset spans the demonstrated speed range with a matching median value of $r$. Evaluation reuses the main evaluation's initial-condition draws at each speed. \Cref{tab:app_scaling} reports the values corresponding to the gray curves in \Cref{fig:capability}. Nominal success increases from 0.27 to 0.44 to 1.00 as the training set grows from 50 to 100 to 297 demonstrations. At $r=2$, the corresponding values are 0.46, 0.44, and 0.53, with overlapping Wilson intervals. The tested subsets therefore show a strong association between demonstration count and nominal success but not between demonstration count and success at $r=2$.

\begin{table}[htb]
\centering
\small
\caption{Task success of ACT-A policies trained with different numbers of demonstrations, 100 episodes per cell with Wilson $95\%$ intervals, under identical evaluation draws across policies. Mint highlights the full 297-demonstration policy used for the main ACT-A results.}
\label{tab:app_scaling}
\begin{tabular}{@{}lccc@{}}
\toprule
Condition & $r=1$ & $r=1.5$ & $r=2$ \\
\midrule
ACT-A ($n{=}50$) & 0.27 [0.19, 0.36] & 0.23 [0.16, 0.32] & 0.46 [0.37, 0.56] \\
ACT-A ($n{=}100$) & 0.44 [0.35, 0.54] & 0.65 [0.55, 0.74] & 0.44 [0.35, 0.54] \\
\rowcolor{oursmint} ACT-A ($n{=}297$) & 1.00 [0.96, 1.00] & 0.91 [0.84, 0.95] & 0.53 [0.43, 0.62] \\
\bottomrule
\end{tabular}
\end{table}

\section{Hand--Parcel Relative Motion}\label{app:motion}
\begin{figure}[htb]
\centering
\includegraphics[width=0.485\textwidth]{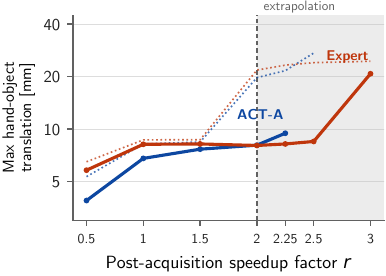}
\includegraphics[width=0.485\textwidth]{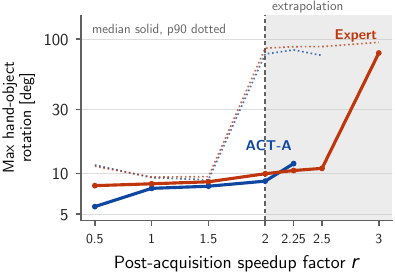}
\caption{\textbf{Hand-Parcel Relative Motion Across Execution Speeds.} Median and 90th-percentile peak hand--parcel relative translation (left) and rotation (right) for the expert and ACT-A. Translation and rotation remain limited across the demonstrated range $r\le2$, despite the substantial expert--learner difference in task success at $r=2$. The shaded region denotes execution speeds outside the demonstrated range.}
\label{fig:app_motion}
\end{figure}

\Cref{fig:app_motion} compares peak hand--parcel relative translation and rotation for the expert and ACT-A across execution speeds. We focus on these two policies because both achieve 100\% task success at nominal speed but differ substantially at higher execution speeds. These continuous measurements complement the binary retention results in \Cref{ssec:handoffresults}.

\section{Baselines Over the Full Speed Grid}\label{app:baselines}
\begin{figure}[htb]
\centering
\includegraphics[width=0.92\textwidth]{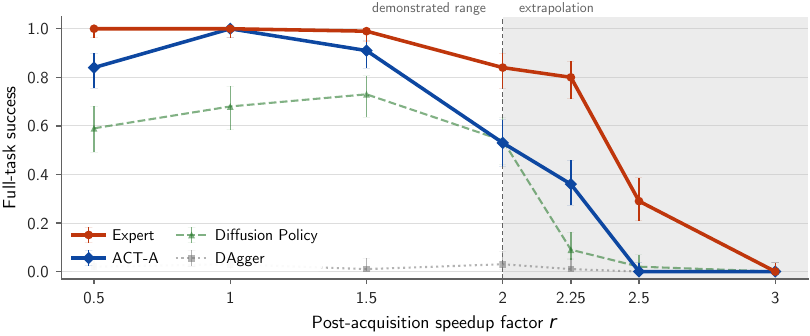}
\caption{\textbf{Baseline Task Success Across the Full Speed Grid.} Task success against the speedup factor for the phases after acquisition for the expert, ACT-A, Diffusion Policy, and DAgger, with Wilson $95\%$ intervals over 100 episodes per point. The shaded band marks extrapolation beyond the demonstrated range, $r>2$.}
\label{fig:app_baselines}
\end{figure}

\Cref{fig:app_baselines} plots the expert, Diffusion Policy, and DAgger across the full evaluation grid. Neither learner reaches expert success at $r=1$, so their expert--learner differences at higher speeds include the nominal success difference. DAgger completes at most 3 of 100 episodes at any evaluated speed, so sensitivity to execution speed cannot be distinguished at this success level.

\section{Out-of-Sample Evaluation of the Realized Contact Margin}\label{app:oos}
\Cref{tab:app_oos} reports the held-out analysis behind \Cref{ssec:certificate}. For each holdout protocol, we fit the Youden threshold and one-dimensional logistic model on the fit population alone. The fitted threshold $\tau^\ast$ lies within $10^{-3}$ of the a priori force closure threshold zero in every protocol. All 414 acquisitions without force closure across the six evaluated policies and speedup factors fail, as do all 79 such acquisitions from the two checkpoints trained with different numbers of demonstrations.

\begin{table}[htb]
\centering
\small
\setlength{\tabcolsep}{4pt}
\caption{Out-of-sample transfer of the realized contact margin as a success predictor. Fit and test populations never overlap. Here, "ba" denotes \emph{balanced accuracy} at the fitted Youden threshold $\tau^\ast$ (ba@$\tau^\ast$) and at the a priori threshold $0$ (ba@$0$), and the Brier score of the fit-set logistic model is paired with the base-rate Brier score of predicting the fit-set success rate. Mint highlights the strongest AUROC values among the acquisition-margin policy and initialization holdouts.}
\label{tab:app_oos}
\begin{tabular}{@{}llrrccccc@{}}
\toprule
Protocol & Margin & $n_{\mathrm{fit}}$ & $n_{\mathrm{test}}$ & AUROC & ba@$\tau^\ast$ & ba@$0$ & Brier & Brier$_0$ \\
\midrule
Speed holdout & $\FC$ (acquisition) & 1089 & 1132 & 0.62 & 0.65 & 0.63 & 0.31 & 0.38 \\
Speed holdout & $\FC$ (reorientation) & 1089 & 1132 & 0.68 & 0.68 & 0.66 & 0.33 & 0.38 \\
Speed holdout & $\FC$ (insertion start) & 1089 & 1132 & 0.72 & 0.71 & 0.62 & 0.38 & 0.38 \\
Speed holdout & $\FC^{(\beta)}$ (acquisition) & 1089 & 1132 & 0.48 & 0.50 & 0.48 & 0.44 & 0.38 \\
Policy holdout (expert) & $\FC$ (acquisition) & 1521 & 700 & 0.70 & 0.54 & 0.50 & 0.25 & 0.29 \\
Policy holdout (DAgger) & $\FC$ (acquisition) & 1846 & 375 & 0.72 & 0.62 & 0.58 & 0.54 & 0.36 \\
Policy holdout (Diffusion Policy) & $\FC$ (acquisition) & 1617 & 604 & 0.73 & 0.74 & 0.72 & 0.18 & 0.26 \\
\rowcolor{oursmint} Policy holdout (ACT-A) & $\FC$ (acquisition) & 1679 & 542 & 0.83 & 0.71 & 0.62 & 0.22 & 0.27 \\
Unseen init.\ (ACT-B) & $\FC$ (acquisition) & 2221 & 298 & 0.70 & 0.68 & 0.74 & 0.18 & 0.25 \\
\rowcolor{oursmint} Unseen init.\ (ACT-C) & $\FC$ (acquisition) & 2221 & 300 & 0.83 & 0.79 & 0.78 & 0.16 & 0.25 \\
Expert in-dist.\ to learners extrap. & $\FC$ (acquisition) & 300 & 732 & 0.65 & 0.54 & 0.67 & 0.51 & 0.78 \\
\bottomrule
\end{tabular}
\end{table}

\paragraph{Continuous Margin Conditioned on Execution Speed.}
To separate the sign-based force closure test from the information carried by the positive margin magnitude, we restrict this analysis to acquisitions with $\FC>0$ and condition on $r\in\{1,1.5,2\}$. Within this subset, the association between $\FC$ and task success varies across policies. In logistic models conditioned on $r$, the standardized coefficient on $\FC$ is $0.24$ with $95\%$ interval $[-0.28,0.75]$ for the expert, $0.61$ $[0.09,1.12]$ for ACT-A, $-0.62$ $[-0.98,-0.26]$ for ACT-B, and $-0.46$ $[-0.87,-0.05]$ for ACT-C; the pooled ACT model with initialization fixed effects gives $-0.21$ $[-0.43,0.00]$. Thus, after conditioning on execution speed and restricting the analysis to grasps with force closure, larger positive margins do not exhibit a uniform association with task success across the evaluated policies. This is distinct from the sign-based result: all 173 acquisitions with $\FC\le0$ in the same speed range fail, consistent with the broader 414-episode finding of \Cref{ssec:certificate}.

\paragraph{Risk-Adjusted Margin with Fixed Friction.}
We also evaluate the FIRMGrasp risk-adjusted margin $\FC^{(\beta)}$ at $\beta=0.95$ under a Gaussian friction prior with standard deviation $0.15$ \cite{enwerem2026firmgrasp}, reported in the fourth row of \Cref{tab:app_oos}. This margin encodes sensitivity to friction uncertainty, whereas ParcelStow fixes friction at $\mu=0.5$ and varies execution timing. Under this evaluation with fixed friction, $\FC^{(\beta)}$ is nonpositive for essentially every acquired episode and has held-out AUROC $0.48$. These measurements characterize the behavior of the risk-adjusted margin under ParcelStow's setting with fixed friction rather than its performance under variation in friction.

\section{Reproducibility and Benchmark Components}\label{app:release}
ParcelStow version \texttt{v1.0.0} is available at \href{\parcelstowrepotxt}{\color{black}\nolinkurl{\parcelstowrepotxt}}. The release includes episode records and scripts for recomputing the quantitative summaries in the main paper; \Cref{tab:app_release} lists its components. A new state-based policy can be evaluated by implementing the observation and action interfaces of \Cref{ssec:stactctrl}.

\begin{table}[htb]
\centering
\small
\caption{Components of the ParcelStow Release.}
\label{tab:app_release}
\begin{tabular}{@{}p{0.33\textwidth}p{0.60\textwidth}@{}}
\toprule
Component & Purpose \\
\midrule
Isaac Lab ParcelStow task & Reproduce the acquisition, reorientation, insertion, and release task with variation in execution speed. \\
Geometry and trajectory files specified before evaluation & Fix the parcel, the receptacle, the start distribution, the task path, and the declared speedup grid. \\
Scripted reference expert & Establish the reference task success rate across execution speeds and collect demonstrations. \\
DAgger, Diffusion Policy, and ACT drivers & Supply reference imitation baselines under a common interface. \\
Evaluation and handoff tools & Measure task success rates, stage outcomes, hand--parcel relative motion, and relative-motion handoffs. \\
Tests of physical integrity & Exercise dynamics of the free parcel, collisions, success predicates, phase timing, and non-mutating diagnostics. \\
Demonstrations and episode records & Support training and matched comparisons between episodes under the fixed protocol. \\
Analysis and reproduction scripts & Recompute the reported quantitative summaries from the released episode records and reproduce the principal evaluation plots. \\
\bottomrule
\end{tabular}
\end{table}

\end{document}